\documentclass[pdflatex,sn-mathphys-num]{sn-jnl}

\usepackage{booktabs}
\usepackage{subcaption}

\usepackage{graphicx}%
\usepackage{multirow}%
\usepackage{amsmath,amssymb,amsfonts}%
\usepackage{amsthm}%
\usepackage{mathrsfs}%
\usepackage[title]{appendix}%
\usepackage{xcolor}%
\usepackage{textcomp}%
\usepackage{manyfoot}%
\usepackage{booktabs}%
\usepackage{algorithm}%
\usepackage{algorithmicx}%
\usepackage{algpseudocode}%
\usepackage{listings}%

\usepackage{caption}
\usepackage{subcaption}
\theoremstyle{thmstyleone}%
\theoremstyle{thmstyletwo}%

\theoremstyle{thmstylethree}%
\usepackage{booktabs}
\usepackage{pifont}
\newcommand{\cmark}{\ding{51}}
\newcommand{\xmark}{\ding{55}}%

\begin{document}

\title[Article Title]{EchoAgent: Guideline-Centric Reasoning Agent for Echocardiography Measurement and Interpretation}

\author*[1]{\fnm{Matin} \sur{Daghyani}}\email{matin.daghyani@ece.ubc.ca}
\equalcont{These authors contributed equally to this work.}

\author*[1]{\fnm{Lyuyang} \sur{Wang}}\email{anne@ece.ubc.ca}
\equalcont{These authors contributed equally to this work.}

\author[1]{\fnm{Nima} \sur{Hashemi}}

\author[1]{\fnm{Bassant} \sur{Medhat}}

\author[1]{\fnm{Baraa} \sur{Abdelsamad}}

\author[1]{\fnm{Eros} \sur{Rojas Velez}}

\author[1]{\fnm{XiaoXiao} \sur{Li}}

\author[2]{\fnm{Michael Y. C.} \sur{Tsang}}

\author[1,2]{\fnm{Christina} \sur{Luong}}

\author[1,2]{\fnm{Teresa S.M.} \sur{Tsang}}

\author[1]{\fnm{Purang} \sur{Abolmaesumi}}

\affil*[1]{\orgname{University of British Columbia}, \orgaddress{\city{Vancouver}, \state{British Columbia}, \country{Canada}}}

\affil[2]{\orgname{Vancouver General Hospital}, \orgaddress{\city{Vancouver}, \state{British Columbia}, \country{Canada}}\footnote{T. S.M. Tsang and P. Abolmaesumi are joint senior authors.}}


\abstract{
\textbf{Purpose:} Echocardiographic interpretation requires video-level reasoning and guideline-based measurement analysis, which current deep learning models for cardiac ultrasound do not support. We present EchoAgent, a framework that enables structured, interpretable automation for this domain. \textbf{Methods:} EchoAgent orchestrates specialized vision tools under Large Language Model (LLM) control to perform temporal localization, spatial measurement, and clinical interpretation. A key contribution is a measurement-feasibility prediction model that determines whether anatomical structures are reliably measurable in each frame, enabling autonomous tool selection. We curated a benchmark of diverse, clinically validated video--query pairs for evaluation. \textbf{Results:} EchoAgent achieves accurate, interpretable results despite added complexity of spatiotemporal video analysis. Outputs are grounded in visual evidence and clinical guidelines, supporting transparency and traceability. \textbf{Conclusion:} This work demonstrates the feasibility of agentic, guideline-aligned reasoning for echocardiographic video analysis, enabled by task-specific tools and full video-level automation. EchoAgent sets a new direction for trustworthy AI in cardiac ultrasound. Our code will be made publicly available at \url{https://github.com/DeepRCL/EchoAgent}.
}




\maketitle

\section{Introduction}\label{sec1}

Echocardiography is a central imaging modality in cardiovascular care, providing non-invasive, real-time assessment of cardiac structure and function~\cite{lang2015recommendations, keller2023echocardiography}. Point-of-care-ultrasound (POCUS) extends this capability to the bedside, where focused examinations by non-specialists address targeted clinical questions. In routine practice, these focused scans capture only a fraction of the data acquired in comprehensive studies (10–20 \emph{vs.} \ $>100$ cine clips), shifting the burden to clinicians to decide which measurements are feasible and diagnostically sufficient under time pressure. The result is high variability, knowledge-intensive decision making, and frequent uncertainty about when to stop scanning, what to measure next, and how to integrate partial evidence into a safe conclusion. These are precisely the situations where structured reasoning, not only perception, becomes critical for usability.

Artificial intelligence (AI) has shown promise in echocardiographic analysis, particularly through general-purpose foundation models such as EchoPrime~\cite{vukadinovic2024echoprime} and EchoApex~\cite{amadou2024echoapex}. These models leverage large-scale vision–language or self-supervised training on millions of echocardiographic videos or images, enabling capabilities such as view classification, segmentation, and disease detection. Within POCUS, however, real-world interpretation requires more than static predictions: the system must judge view adequacy, infer which quantitative measurements are feasible, and justify recommendations in language aligned with clinical standards. A unified framework that connects quantitative analysis with guideline-based interpretation, and that exposes intermediate reasoning and visual evidence, could therefore improve both accuracy and trust in POCUS workflows.

Agentic systems allow language language models to orchestrate specialized tools. MedRAX~\cite{pmlr-v267-fallahpour25a} demonstrated tool-based reasoning for chest radiographs, but its assumptions do not hold for echocardiography. Unlike static images, echo requires integrated spatialtemporal reasoning and continuous quality control. An agent for echo must align each step with clinical protocols while checking its reasoning against echocardiography guidelines and issuing safe-to-answer decisions. Such an agent closes the loop between perception and reasoning, reducing hallucinations and making bedside recommendations auditable. 

To address these limitations, we introduce \textbf{EchoAgent}, a guideline-centric agentic framework for unified measurement analysis and interpretation in echo with an emphasis on POCUS usability. EchoAgent operates directly on video input and integrates domain specific tools under Large Language Model (LLM) orchestration. Its main contributions are:
\begin{enumerate}
\item \textbf{Measurement-feasibility forecasting and planning} to determine whether required anatomical structures are present and measurable, reducing cognitive load in focused POCUS exams. 
\item \textbf{Interpretable, guideline-aligned reasoning}, with each step grounded in visual evidence and verified against echocardiography standards; the agent exposes its intermediate rationale, highlights supporting echo cine-frame evidence, and issues “safe-to-answer” or escalation recommendations that clinicians can audit. 
\item \textbf{End-to-end video level analysis} that accounts for temporal dynamics and variable image quality, essential for real-world POCUS where operator skill and patient factors vary.
\end{enumerate}
By elevating reasoning to deciding what is feasible, and why the answer is trustworthy, EchoAgent is designed to make POCUS more usable for non-specialists while preserving the rigor of guideline based echocardiography.

\section{Related Work}\label{sec2}
\subsection{Foundation Models for Echocardiography}
Recent work has explored large-scale echocardiography models trained using self-supervised or contrastive objectives. Vision-only foundation models focus on representation learning for downstream tasks but cannot reason over textual input or provide interpretable outputs~\cite{amadou2024echoapex}. Vision–language models (VLMs) trained on video–report pairs have demonstrated impressive zero-shot generalization for classification~\cite{vukadinovic2024echoprime}, but their predictions lack step-by-step justification or alignment with clinical guidelines. As a result, outputs cannot be verified or audited, and such systems do not support interactive visual question answering or structured measurement interpretation. VLM-based approaches~\cite{QinYi_MultiAgent_MICCAI2025, she2025echovlm} further require task-specific fine-tuning on image–text pairs, which are scarce in echocardiography, and limit evaluation to static frames rather than full video sequences. Some recent work has introduced automated measurement prediction in echo videos~\cite{sahashi2025artificial}, but these models do not perform multi-step reasoning or integrate interpretive standards. In contrast, our framework offers both numerical and interpretive outputs, grounded in visual evidence and expert-derived guidelines.

\subsection{Agentic AI in Medicine}
Agentic frameworks enable large language models to plan and invoke external tools for clinical reasoning~\cite{yao2023react, schick2023toolformer, shen2023hugginggpt, kim2024mdagents, fathi2025aura, bani2024magda, pmlr-v267-fallahpour25a}. Recent adaptations incorporate multi-agent and planner–executor designs to improve retrieval-augmented reasoning and reduce hallucinations in medical tasks~\cite{wind2025agentic_radiology, masayoshi2025ehrmcp}. However, these systems target static images or structured data and do not address the spatiotemporal complexity of echocardiography. EchoAgent extends this paradigm by supporting full video-level automation with interpretable, guideline-grounded outputs.

\section{EchoAgent}\label{sec3}
\subsection{Overview}

EchoAgent is a modular framework designed for end-to-end echocardiographic measurement and interpretation through iterative reasoning. It comprises three primary components: a set of domain-specific tools $\mathcal{T}$ for video analysis, measurement extraction, and guideline lookup; an indexed guideline database $\mathcal{D}$; and a large language model $\mathcal{M}$ that orchestrates the overall process. Given a video $V$ and a user query $Q$, the agent incrementally determines its next step using prior observations. At each step, it selects a tool from $\mathcal{T}$, formulates its input parameters, and updates its internal history with the resulting output. This iterative process continues until the agent concludes that it has sufficient information to produce a guideline-consistent response. The following subsections detail the agentic decision process and describe the specialized tools integrated into the framework.

\subsection{Agentic Design}

Given an echocardiogram $V$ and user query $Q$, the agent’s goal is to iteratively gather the information needed to answer the query in a guideline-consistent manner. It integrates two information sources: (i) the video $V$, from which clinically valid measurements, key frames, and feasibility assessments are extracted; and (ii) the guideline database $\mathcal{D}$, which provides structured clinical knowledge for interpreting results.

Instead of a fixed pipeline, the language model orchestrator $\mathcal{M}$ operates adaptively in an iterative loop~\cite{yao2023react} of three phases: \textit{observation} (incorporating new measurements or retrieved data), \textit{thought} (identifying remaining uncertainties), and \textit{action} (selecting a tool from $\mathcal{T}$ and producing input parameters). Each action’s output is fed back into the loop to guide subsequent reasoning.

The language model does not process $V$ directly but interfaces with modular tools, such as a feasibility prediction model, measurement segmentation models, and a cardiac phase estimator that return structured outputs. In parallel, it may employ a clinical context retrieval tool to access relevant information from $\mathcal{D}$, such as threshold values, diagnostic criteria, or interpretive rules grounded in clinical guidelines.

The process continues until the agent determines it has sufficient information to answer the query, invoking a \texttt{FINISH} action. The final output includes both the answer and a structured trace of reasoning steps, ensuring transparency and verifiability. The complete reasoning loop is formalized in Algorithm~\ref{alg:echoagent}.

\begin{figure}[t]
    \centering
    \includegraphics[width=\linewidth]{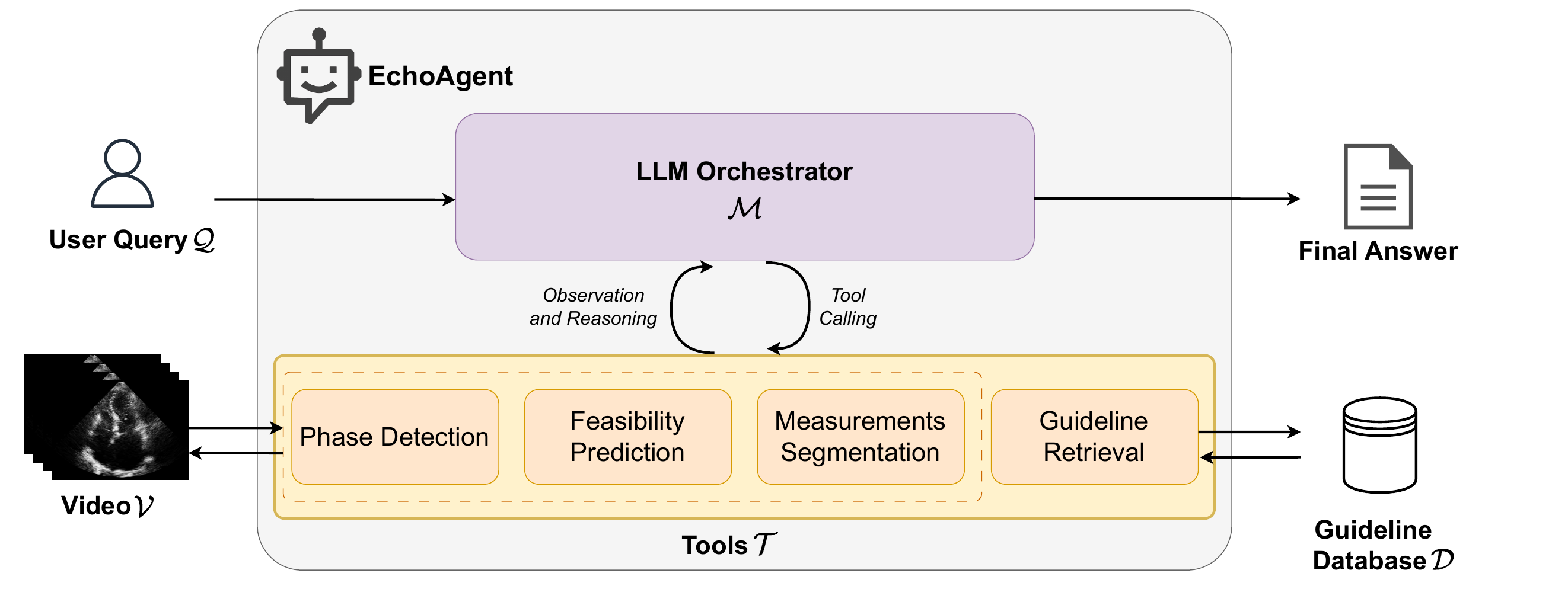}
    \caption{
    \textbf{Overview of the EchoAgent framework.}
    EchoAgent is a guideline-centric reasoning system that integrates specialized visual tools under the orchestration of an LLM. 
    Given a clinician’s natural-language query and an echo video, the LLM orchestrator dynamically invokes tools for phase detection, feasibility prediction, measurement segmentation, and guideline retrieval to produce interpretable, guideline-aligned answers. 
    The system iteratively reasons over intermediate observations to ensure that all measurements are visually grounded and clinically validated.
    }
    \label{fig:framework}
\end{figure}
\begin{algorithm}[t]
\caption{EchoAgent: Iterative Agentic Reasoning Loop}
\label{alg:echoagent}
\begin{algorithmic}[1]
\Require Query $Q$, video $V$, tool set $\mathcal{T}$, guideline database $\mathcal{D}$, language model $\mathcal{M}$
\Ensure Final response $R$

\State Initialize history $H \gets \varnothing$
\State Set step counter $i \gets 0$, maximum steps $K$

\While{$i < K$}
    \State $\mathcal{O}_i \gets \mathcal{M}.\texttt{observe}(Q, H)$
    \State $(T_i, P_i) \gets \mathcal{M}.\texttt{reason}(\mathcal{O}_i, \mathcal{T})$

    \If{$T_i = \texttt{FINISH}$}
        \State $R \gets \mathcal{M}.\texttt{generateAnswer}(H)$
        \State \Return $R$
    \ElsIf{$T_i = \texttt{search\_guideline}$}
        \State $Y_i \gets \texttt{executeTool}(T_i, P_i, \mathcal{D})$
    \Else
        \State $Y_i \gets \texttt{executeTool}(T_i, P_i, V)$
    \EndIf

    \State $H \gets H \cup \{(T_i, P_i, Y_i)\}$
    \State $i \gets i + 1$
\EndWhile

\State \Return $\mathcal{M}.\texttt{generateAnswer}(H)$
\end{algorithmic}
\end{algorithm}
\subsection{Tools Integration}
To support the LLM's reasoning, EchoAgent integrates specialized tools that extract clinically meaningful information from echocardiographic videos and standard guidelines. The vision tools include a cardiac phase detector for identifying end‑diastolic (ED) and end‑systolic (ES) frames, a measurement‑feasibility model for assessing anatomical suitability, and segmentation models for standard B‑mode linear measurements. In addition, a guideline‑retrieval tool enables the agent to query standard references to interpret user queries and evaluate the clinical relevance of predicted values. 

\textbf{Cardiac Phase Detection Tool:}
We trained this module in a self-supervised manner~\cite{dezaki2021echo} to learn frame-level representations that capture temporal and morphological variations across the cardiac cycle. Since echocardiographic measurements are inherently frame-specific, identifying the end-diastolic (ED) and end-systolic (ES) frames is crucial. These key frames define physiologically meaningful reference points corresponding to maximal and minimal ventricular volumes, respectively. Accurate detection of ED and ES frames enables consistent measurement of parameters such as chamber dimensions, wall thickness, and strain, which are essential for reliable cardiac function assessment. After completing the self-supervised training, we froze the encoder and appended a lightweight classifier to identify the ED and ES frames. Given a video sequence \( x \in \mathbb{R}^{T \times H \times W} \) with \(T\) frames, the classifier takes the embeddings of all frames as input and outputs the indices corresponding to the ED and ES frames.

\textbf{Feasibility Prediction Tool:}
Although standard echocardiographic views are defined for specific quantitative measurements~\cite{mitchell2019guidelines}, not all videos labeled as a given view are suitable for every measurement associated with that view. View classification alone does not guarantee measurement validity, as variability in zoom level, image quality, and anatomical visibility often makes certain measurements unreliable even when the view label is correct.
Moreover, in real-world usage, user queries may not always specify what to measure, placing the burden on the agent to infer what is visually feasible and clinically relevant. To address this, we introduce a feasibility prediction model that identifies which measurements are supported by the image content.

Given a frame from an echocardiogram $x \in \mathbb{R}^{H \times W}$, the model predicts a binary vector $\hat{y} \in \{0,1\}^m$, where $m$ is the number of linear measurements and $\hat{y}_j = 1$ indicates measurement $j$ is feasible. The model is trained on key frames (end-diastole or end-systole) annotated with which measurements were performed originally by sonographers. We construct a binary label vector $y$ per frame and train a ResNet-50 with a sigmoid output head using binary cross-entropy loss, framing the task as multi-label classification. This tool allows the agent to infer what is measurable—even without an explicit user request—and avoid unreliable predictions, improving both robustness and autonomy in measurement selection.

\textbf{Linear Measurement Tool:}
We employ EchoNet-Measurements~\cite{sahashi2025artificial} as the core tool for computing echocardiographic linear measurements. After phase detection and feasibility assessment, the candidate frame and corresponding measurement name are provided to the tool, which applies dedicated pre-trained weights for each measurement type. The tool outputs both the measurement localization and quantitative value, which are retrieved by the agent for subsequent reasoning.

\textbf{Clinical Context Retrieval Tool:}
LLMs are prone to hallucination, particularly in echocardiography, where limited data and guideline-based reasoning demand precise numerical interpretation. To ensure reliable outputs, EchoAgent follows clinical standards by using a retrieval tool that accesses guideline documents during reasoning. Unlike standard retrieval-augmented generation (RAG)~\cite{lewis2020retrieval}, which performs a single initial retrieval, the agent dynamically retrieves relevant guidelines whenever additional context is needed. The tool employs a Fossil-based dense index over curated guideline text~\cite{xiong2024benchmarking} to return top-K relevant passages, providing clinically grounded context that informs the agent’s decisions.

\begin{figure}[t]
  \centering
  \begin{subfigure}[t]{0.49\linewidth}
    \centering
    \includegraphics[page=2, width=\linewidth]{figures/echoagent-example.pdf}
    \label{fig:example-left}
  \end{subfigure}
  \hfill
  \begin{subfigure}[t]{0.49\linewidth}
    \centering
    \includegraphics[page=1, width=\linewidth]{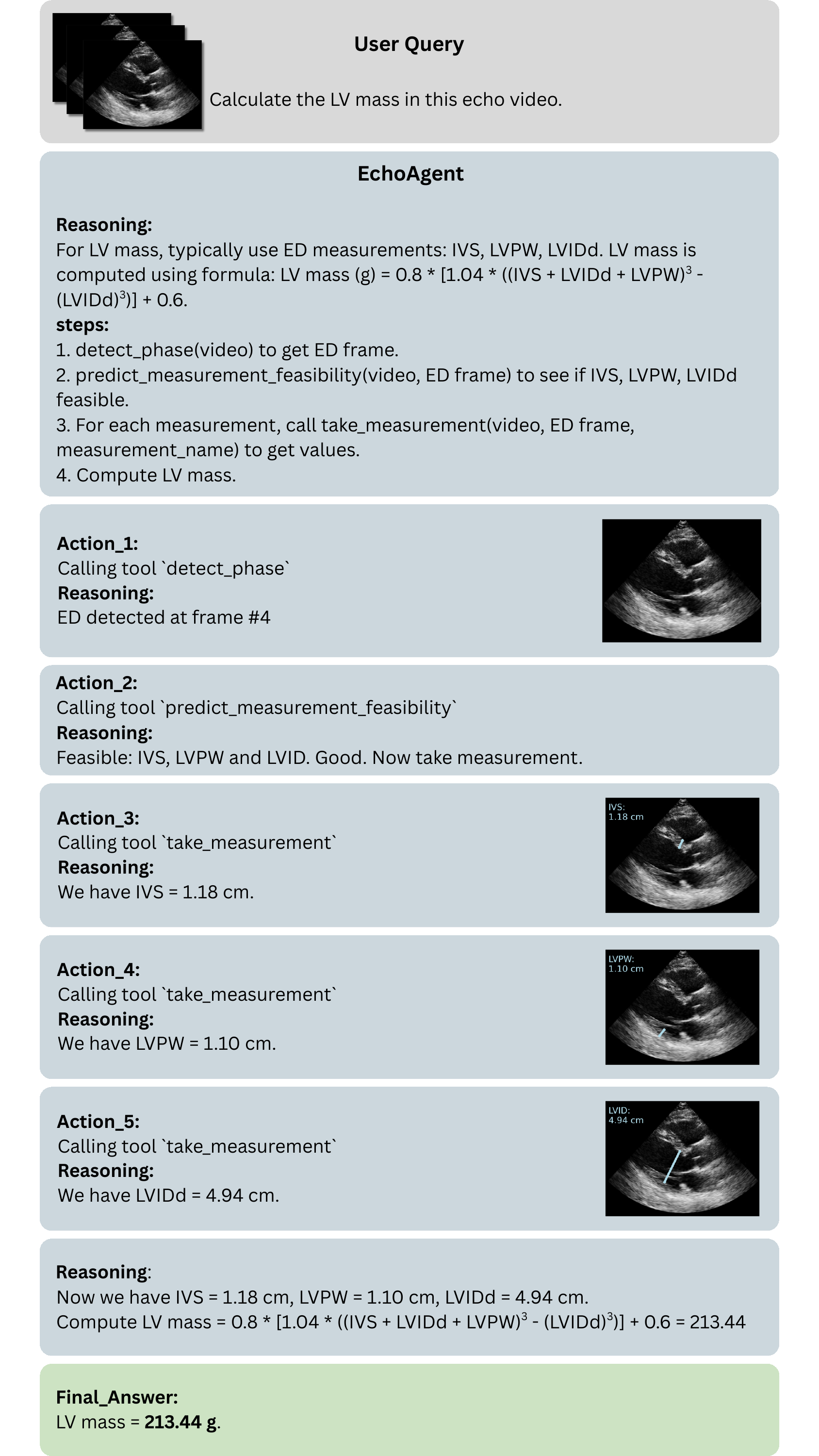}
    \label{fig:example-right}
  \end{subfigure}
  \caption{\textbf{EchoAgent Interaction Flow.} Examples with visualizations of how EchoAgent handles a user query through reasoning and interaction with different tools. The example on the left is an easier case; the example on the right requires derived calculations.}
  \label{fig:EchoAgent_interact_flow}
\end{figure}

\section{Evaluation}\label{sec4}
\subsection{Dataset}
The dataset used in this study was sourced from the echocardiography database at Vancouver General Hospital, with ethics approval. It includes 207{,}869 videos from 56{,}218 exams across 13 standard views, and 305{,}229 annotated frames spanning 16 measurement types. All 16 were used to train the measurement feasibility model. However, only seven standard linear measurements that overlap between our private dataset and the EchoNet-Measurements dataset were used to evaluate the full framework. The data were split into training, validation, and test sets, with no patient or video overlap. 

For end-to-end evaluation, we curated a benchmark of 60 representative test examples, each consisting of a video, a clinical question, and its corresponding answer, covering diverse views, image qualities, and measurement types. Each was refined and verified by an expert sonographer, with answers grounded in clinical reports. The sonographer also assigned difficulty levels: \textit{easy} (single measurement), \textit{medium} (multiple measurements), and \textit{difficult} (requiring reasoning or derived calculations), enabling more granular evaluation of the agent’s capabilities.

\subsection{Implementation Details}
EchoAgent employs GPT-OSS-20B~\cite{openai2025gptoss120bgptoss20bmodel} as its primary LLM backbone. This open-source model, released by OpenAI, is fine-tuned for structured reasoning and tool use. The framework is modular with respect to the LLM orchestrator. We also evaluate Qwen3Coder~\cite{qwen3technicalreport} and LLaMA 3.1~\cite{dubey2024llama}, both optimized for tool-augmented inference. The agent interacts with tools through structured JSON calls specifying function names and arguments. All models use their official implementations and pretrained weights. Input videos are resized to $224\times224$ for feasibility and phase prediction, while measurement models operate at $480\times640$ to preserve spatial fidelity. Each query is limited to 15 reasoning iterations within the tool-use loop. All experiments were conducted on a single NVIDIA B200 GPU (180 GB VRAM).






\begin{table}[t]
\centering
\caption{Accuracy and failure-case analysis of LLMs on the benchmark (n=60), evaluated on final answers. Accuracy is reported across difficulty levels. Failure cases include \textbf{Tool Calling} errors (hallucinated or invalid tool calls), and \textbf{Final Conclusion} errors (clinically invalid interpretations despite correct evidence). Tool-related measurement errors (e.g., incorrect values or phase detection) are excluded to isolate LLM-orchestration weaknesses.}
\label{tab:model_results}
\begin{tabular}{@{}lcccccc@{}}
\toprule
\multirow{2}{*}{\textbf{Model}} 
& \multicolumn{4}{c}{\textbf{Accuracy}} 
& \multicolumn{2}{c}{\textbf{\#Failure Cases}} \\
\cmidrule(lr){2-5} \cmidrule(lr){6-7}
& Easy & Medium & Difficult & Overall & Tool Calling   & Final Conclusion  \\
\midrule
LLaMA 3.1 (8B)~\cite{dubey2024llama}     & 0.33 & 0.29 & 0.00 & 0.28 & 17 & 22 \\
Qwen3Coder (30B)~\cite{qwen3technicalreport}       & 0.44 & 0.29 & 0.38 & 0.42 & 2 & 24 \\
GPT-OSS (20B)~\cite{openai2025gptoss120bgptoss20bmodel} & \textbf{0.64} & \textbf{0.57} & \textbf{0.50} & \textbf{0.62} & 2 & 11 \\
\bottomrule
\end{tabular}
\end{table}
\subsection{EchoAgent Evaluation}
We evaluate the impact of different LLMs within our agentic framework by replacing the underlying language model and measuring overall accuracy. Correctness is determined by a GPT-4o-based judge. As shown in Table~\ref{tab:model_results}, GPT-OSS achieves the highest overall accuracy of 0.62, with consistently strong results across all difficulty levels. Notably, all models perform best on Easy queries, while struggling more with Medium and especially Difficult ones, indicating a growing challenge in handling more complex reasoning and decision-making steps within the agent workflow.

To better understand failure modes, we analyze the types of incorrect final answers across LLMs (see Table~\ref{tab:model_results}). One failure type involves tool calls with hallucinated or invalid arguments, reflecting poor control over structured actions. Another stems from drawing clinically invalid conclusions despite having relevant evidence, revealing weaknesses in reasoning. LLaMA 3.1 fails more often due to the former, while Qwen3Coder shows more of the latter. These contrasting patterns suggest that some models struggle with tool reliability, while others falter in inference. GPT-OSS avoids both dominant failure types, indicating a stronger integration of tool use and clinical reasoning within the agentic loop.

\subsection{Tools Evaluation}

\begin{table*}[t]
\centering
\caption{Tool-specific results on ED/ES frames of the test set.}
\label{tab:tools_results}
\vspace{6pt}

\begin{subtable}[t]{0.32\textwidth}
\centering
\caption{Linear-measurement tool}
\label{tab:linear}
\vspace{3pt}
\begin{tabular}{lc}
\toprule
Measurement & MAE\\
& (cm) \\
\midrule
IVS & 0.13 \\
LVID & 0.31 \\
LVPW & 0.22 \\
LA & 0.29 \\
Aorta & 0.28 \\
Aortic root & 0.27 \\
RV base & 0.28 \\
\bottomrule
\end{tabular}
\end{subtable}
\hfill
\begin{subtable}[t]{0.64\textwidth}
\centering
\caption{Feasibility-prediction tool}
\label{tab:feasibility}
\vspace{3pt}
\begin{tabular*}{0.95\textwidth}{@{\extracolsep{\fill}}lccc@{}}
\toprule
 & Prec. & Rec. & F1 \\
\midrule
Micro & 0.84 & 0.87 & 0.86 \\
Macro & 0.85 & 0.86 & 0.85 \\
\bottomrule
\end{tabular*}

\vspace{-1pt} 

\caption*{(c) Phase-detection tool}
\hypertarget{tab:phase}{}
\vspace{1pt}
\begin{tabular}{lc}
\toprule
\multicolumn{1}{c}{Frame} & MAE (frames)  \\
\midrule
ED & 1.95 (3.05)  \\
ES & 4.25 (6.63)  \\
\bottomrule
\end{tabular}
\end{subtable}
\end{table*}

\noindent
\textbf{Linear Measurement Tool.} We evaluate pre-trained EchoNet‑Measurements models on our internal test set for the seven overlapping linear measurements shared with their original dataset. As shown in Table~\ref{tab:linear}, the models demonstrate strong agreement with sonographer-provided ground truth, with low mean absolute error (MAE) values reported in centimeters. Notably, the reported metrics are comparable to those reported in the original paper, indicating that the models generalize well to our private dataset without additional fine-tuning.

\noindent
\textbf{Feasibility Prediction Tool.} We evaluate the performance of the feasibility prediction model on over 30{,}000 end-diastolic (ED) or end-systolic (ES) frames from the test set, each annotated with at least one measurement (Table~\ref{tab:feasibility}). The model achieves strong frame-level prediction quality, with a micro F1-score of 0.86 and macro F1-score of 0.85, indicating consistent performance across both frequent and rare classes. This level of accuracy enables the agent to reliably identify frames in which the cardiac structures are sufficiently visible and well-positioned for valid measurement extraction, effectively reducing spurious tool calls and improving downstream precision.

\noindent
\textbf{Phase Detection Tool.} We adopt the Mean Absolute Error (MAE) metric from~\cite{YanYin_Latent_MICCAI2025} as the average frame-wise distance between each ground truth (GT) frame and the closest predicted frame. For each video, we evaluate a single cardiac cycle by selecting the most probable ED and ES predictions, consistent with the available annotations. As shown in Table~\hyperlink{tab:phase}{2c}, the model achieves an MAE of 1.95 (3.05) for ED and 4.25 (6.63) for ES, with higher error in ES due to greater visual ambiguity near systole.

\subsection{Ablation Study}

\begin{table}[t]
\centering
\caption{Ablation Study}
\label{tab:ablation-feasible-rag}
\setlength{\tabcolsep}{6pt}
\begin{tabular}{cc|cccc}
\toprule
\multicolumn{2}{c}{Components} & \multicolumn{4}{c}{Accuracy} \\
\cmidrule(lr){1-2}\cmidrule(lr){3-6}
Feasibility Det. & Clinical Context Retrieval & Overall & Easy & Medium & Difficult \\
\midrule
\xmark & \xmark & 0.48 & 0.56 & 0.29 & 0.25 \\  
\xmark & \cmark & 0.52 & 0.56 & 0.43 & 0.38 \\
\cmark & \xmark & 0.53 & 0.58 & 0.29 & \textbf{0.50} \\
\addlinespace[2pt]
\cmark & \cmark & \textbf{0.62} & \textbf{0.64} & \textbf{0.57} & \textbf{0.50} \\
\bottomrule
\end{tabular}
\end{table}

To evaluate the contribution of each component, we progressively add the feasibility detector and the clinical context retrieval tool. As shown in Table~\ref{tab:ablation-feasible-rag}, when both are absent, performance drops notably (0.48 overall), indicating that the agent struggles to reason without structured guidance or measurement validation. Adding the context retrieval tool improves overall and medium-level accuracy, confirming that guideline-derived knowledge reduces reasoning errors. Incorporating the feasibility detector further boosts performance, particularly on difficult cases, by filtering unreliable measurements. The full model achieves the highest accuracy across all levels, demonstrating the complementary value of both components.

\section{Discussion}\label{sec5}
EchoAgent demonstrates that guideline-centric tool orchestration enables accurate, interpretable echocardiographic analysis with end-to-end reasoning consistent with agentic systems in other modalities, such as MedRAX for chest X-ray. By grounding actions in retrieved guidelines and exposing intermediate steps, the framework enhances transparency and reduces unsupported conclusions in point-of-care workflows. However, it remains vulnerable to error propagation, as final decision quality depends on upstream tool accuracy. Moreover, the lack of explicit uncertainty modeling limits calibrated confidence and robust deferral when evidence is insufficient. Future work should incorporate uncertainty-aware aggregation, error detection and recovery within the reasoning loop, and broader tool coverage to mitigate failures from missing or imperfect measurements.

\section{Conclusion}\label{sec6}
EchoAgent introduces a guideline‑centric agent that integrates spatiotemporal video analysis, measurement‑feasibility prediction, and clinically grounded interpretation within a unified, LLM‑orchestrated framework. On a curated benchmark, it achieves accurate, interpretable, and auditable performance, supported by strong tool accuracy and reliable reasoning across difficulty levels, highlighting the effectiveness of structured tool orchestration compared with end‑to‑end modeling. By exposing intermediate steps and guideline references, EchoAgent enhances transparency and ensures consistency with established clinical standards. Future work will expand the toolset (e.g., Doppler, volumetric, and additional linear measurements) and extend reasoning from single‑video analysis to multi‑video, study‑level synthesis for comprehensive, context‑aware clinical reasoning.

\bibliography{sn-bibliography}

\end{document}